\documentclass{article}

\PassOptionsToPackage{numbers, compress}{natbib}

 \usepackage[preprint]{neurips_2026}

\usepackage[utf8]{inputenc} 
\usepackage[T1]{fontenc}    
\usepackage{hyperref}       
\usepackage{url}            
\usepackage{booktabs}       
\usepackage{amsfonts}       
\usepackage{nicefrac}       
\usepackage{microtype}      
\usepackage{xcolor}         
\usepackage{amsmath}
\usepackage{wrapfig}
\usepackage{enumitem}

\definecolor{tabblue}{HTML}{1F77B4}
\definecolor{taborange}{HTML}{FF7F0E}
\definecolor{tabgreen}{HTML}{2CA02C}
\definecolor{tabred}{HTML}{D62728}
\definecolor{tabpurple}{HTML}{9467BD}
\definecolor{tabbrown}{HTML}{8C564B}
\definecolor{tabpink}{HTML}{E377C2}
\definecolor{tabgray}{HTML}{7F7F7F}
\definecolor{tabolive}{HTML}{BCBD22}
\definecolor{tabcyan}{HTML}{17BECF}

\usepackage{rotating} 
\usepackage[most]{tcolorbox}
\newtcolorbox{takeaway}{
  colback=blue!5,
  colframe=blue!5,
  width=1\textwidth,
  boxrule=0pt,
  left=5pt, right=5pt, top=3pt, bottom=3pt, 
  boxsep=0pt, 
}

\title{Cornerstones or Stumbling Blocks? Deciphering the Rock Tokens in On-Policy Distillation}

\author{
Yuxuan Jiang$^{1,*}$,
Runchao Li$^{2,*}$,
Shubhashis Roy Dipta$^{1,*}$,
Dawei Li$^{3}$,
Zhao Yang$^{4,\dagger}$\\
$^{1}$University of Maryland, Baltimore County \quad
$^{2}$Case Western Reserve University \\
$^{3}$Arizona State University \quad
$^{4}$VU Amsterdam\quad
$^{*}$Equal contribution \quad
$^{\dagger}$Corresponding author
}

\begin{document}

\maketitle

\begin{abstract}

While recent work in Reinforcement Learning with Verifiable Rewards (RLVR) has shown that a small subset of critical tokens disproportionately drives reasoning gains, an analogous token-level understanding of On-Policy Distillation (OPD) remains largely unexplored. In this work, we investigate high-loss tokens, a token type that—as the most direct signal of student-teacher mismatch under OPD's per-token KL objective—should progressively diminish as training converges according to existing studies; however, our empirical analysis shows otherwise. Even after OPD training reaches apparent saturation, a substantial subset of tokens continues to exhibit persistently high loss; these tokens, which we term Rock Tokens, can account for up to 18\% of the tokens in generated outputs. Our investigation reveals two startling paradoxes. First, despite their high occurrence frequency providing a disproportionately large share of total gradient norms, Rock Tokens themselves remain stagnant throughout training, resisting teacher-driven corrections. Second, through causal intervention, we find that these tokens provide negligible functional contribution to the model's actual reasoning performance. These findings suggest that a vast amount of optimization bandwidth is spent on structural and discourse residuals that the student model cannot or need not internalize. By deconstructing these dynamics, we demonstrate that strategically bypassing these "stumbling blocks" can significantly streamline the alignment process, challenging the necessity of uniform token weighting and offering a more efficient paradigm for large-scale model distillation. Code is available at: \\
\href{https://github.com/YuxuanJiang1/Rock-Token}{https://github.com/YuxuanJiang1/Rock-Token}
\end{abstract}

\section{Introduction}

On-Policy Distillation (OPD) has established itself as a cornerstone of the modern post-training pipeline for Large Language Models (LLMs)~\cite{agarwal2024policy}. By aligning a student model with a superior teacher through trajectories sampled from its own policy, OPD enables a token level of reasoning refinement that goes beyond static Supervised Fine-Tuning (SFT)~\cite{tan2024large}. This effectiveness has been validated by industrial works such as DeepSeek-V4~\cite{deepseekv4}, MiMo~\cite{xiao2026mimo}, and Qwen-3~\cite{qwen3technicalreport}, where OPD serves as a vital component alongside SFT or Reinforcement Learning with Verifiable Rewards (RLVR) to further squeeze out reasoning performance.


Recent studies in Reinforcement Learning with Verifiable Rewards (RLVR)~\cite{dipta2026pa3,dipta2026ganitllm,guo2025deepseek} have revealed that not all tokens contribute equally to model learning: a small subset of critical tokens, such as high-entropy "forking tokens" at decisive branching points, disproportionately drives reasoning gains~\cite{wang2025beyond}. While this token-level perspective has reshaped our understanding of RLVR dynamics, an analogous investigation in OPD remains largely unexplored, despite OPD's reliance on dense token-level supervision. In this work, we initiate such an investigation by focusing on a token type uniquely meaningful to the OPD paradigm: high-loss tokens. Unlike RLVR, where token importance is inferred from policy entropy or reward signals, OPD provides a direct measure of student-teacher mismatch through per-token KL divergence; high-loss tokens thus explicitly mark the positions where the student's policy diverges most sharply from the teacher's, making them natural candidates for the ``critical correction points'' that drive alignment~\cite{lu2025onpolicydistillation}. 

Intuitively, as the student aligns with the teacher during the OPD process, both the magnitude and cardinality of these high-loss tokens are expected to diminish, eventually leading to an optimization plateau. However, our empirical investigations reveal a striking phenomenon: even after training has reached apparent saturation, a stubborn population of tokens—which we formally define as Rock Tokens—continues to exhibit consistently high loss. Notably, while these tokens represent a small minority of the vocabulary (approximately 6\%), they account for as much as 18\% of the total token frequency in the model's output. This persistence challenges conventional understanding of model convergence and raises a fundamental question: Are Rock Tokens indispensable Pillars for policy alignment, or are they Stumbling Blocks that create unnecessary optimization redundancy?

\begin{figure}[t]
  \centering
  \includegraphics[width=\linewidth]{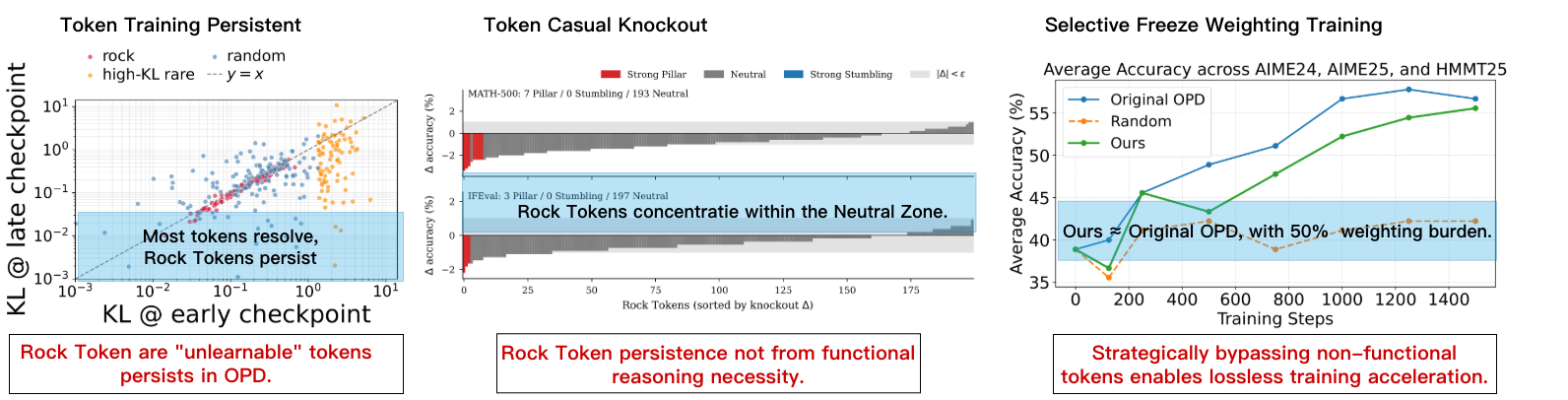}
  \caption{\textbf{The lifecycle and functional impact of Rock Tokens in OPD.} (a) \textbf{Phenomenon}: Identification of optimization-resistant tokens. (b) \textbf{Mechanism}: Causal evidence of structural redundancy via token knock-out. (c) \textbf{Utility}: Performance parity achieved through strategic gradient sparsification.}
  \label{fig:intro}
\end{figure}

To provide a systematic understanding of the Rock Token phenomenon, our research is structured across three investigative dimensions: identifying \textbf{what} are rock tokens, analyzing \textbf{why} they emerge and persist, and uncovering \textbf{how} they collectively impact the overall OPD training process. This comprehensive framework is operationalized through the following three stages:

First, we characterize the identity of Rock Tokens (What). We begin by identifying Rock Tokens linguistic and statistical properties, revealing that they primarily comprise \textbf{syntactic and structural scaffolding}—such as formatting delimiters, whitespace symbols, and high-frequency discourse markers (e.g., ``So'', ``Wait''). To determine the dynamic role of these elements, we employ quantitative gradient dynamics analysis to address \textbf{RQ1: Do Rock Tokens still provide useful learning signals during optimization?} By examining whether their persistent high loss effectively translates into constructive gradients, we assess whether these tokens offer steady directional guidance or merely represent stagnant optimization noise throughout the OPD process.

Second, we investigate the underlying mechanisms behind their persistence (Why). Beyond the actual contribution of Rock Tokens during training is clarified, a more profound question emerges: \textbf{why do these tokens persist in the first place?} We hypothesize that the student model develops a strong decoding dependency on these specific tokens to maintain its reasoning flow, creating a path dependency that resists corrections from the teacher model. This leads to our \textbf{RQ2: Does Rock Token persistence stem from student path dependency?} To investigate this, we conduct \textit{token knock-out} experiments as a form of ablation analysis, examining whether the student's logical coherence is so deeply intertwined with these persistent residuals that it becomes structurally unable to shift its policy toward the teacher's distribution.

Finally, we evaluate how Rock Tokens influence the overall OPD training process (How). Finally, we arrive at a provocative question: what would happen if these tokens were excluded from the very beginning of OPD training? This leads to our final investigation, \textbf{RQ3: What is the genuine functional contribution of Rock Tokens to model training?} By freezing the gradients of these tokens from the onset of training, we intuitively isolate their functional necessity throughout the entire distillation process. This causal intervention allows us to rigorously determine whether Rock Tokens serve as indispensable pillars of the model's reasoning integrity or if they represent a strategic opportunity to bypass computational redundancy and optimize the efficiency of the alignment pipeline.

In summary, our key contributions are as follows:

\begin{itemize}[leftmargin=*,itemsep=1pt]
    \vspace{-2mm}
    \item For the first time, we identify and characterize \textbf{Rock Tokens}, a persistent high-loss phenomenon that unveils a previously hidden aspect of optimization dynamics in On-Policy Distillation.
    \vspace{-1mm}
    \item We deconstruct the intricate dynamics of the OPD process by systematically uncovering the functional roles of Rock Tokens across both the training and inference phases.
    \vspace{-1mm}
    \item Our research reveals a critical inefficiency in current alignment pipelines, demonstrating that a substantial portion of training signals in OPD provides significantly lower-than-expected contributions to the model's ultimate reasoning performance.
\end{itemize}

\section{Rock Tokens} 
\label{ref:recalcitrant_tokens}

\begin{figure}[t]
  \centering
  \includegraphics[width=\linewidth]{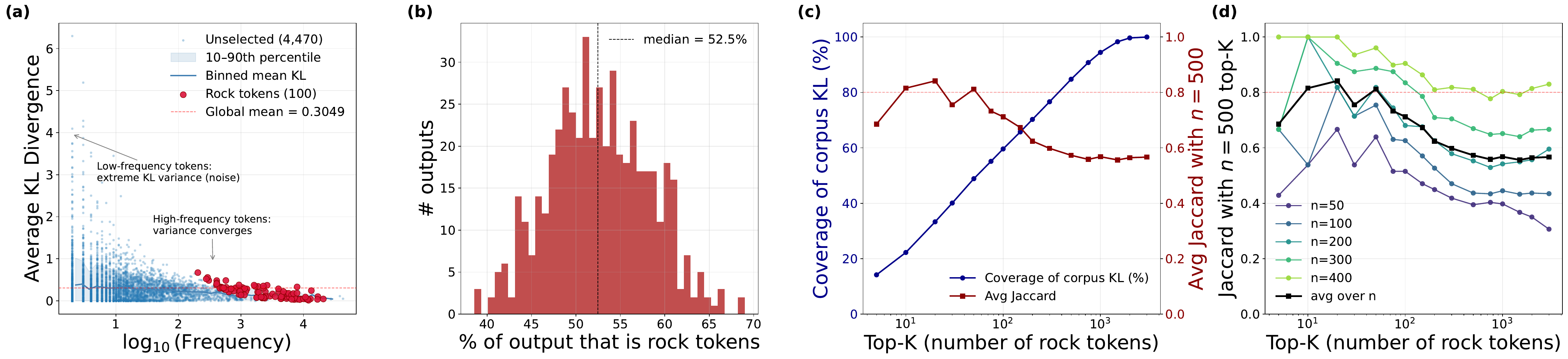}
  \caption{\textbf{Empirical identification and stability of Rock Tokens.}
  \textbf{(a)} Per-token KL $\widehat\ell_v$ vs.\ frequency on $N{=}500$
  MATH-500 trajectories: rare tokens are noise-dominated, while the Rock
  Score $R(v)$ isolates true Rock Tokens (red) at the upper edge of stable
  frequency bands. \textbf{(b)} Per-sequence Rock-Token density
  (median $18.5\%$). \textbf{(c)} Cumulative loss coverage (blue) and
  selection stability (red) vs.\ cutoff $K$; $K{=}100$ balances
  reproducibility with $\approx 60\%$ coverage of the distillation burden.
  \textbf{(d)} Jaccard ranking robustness across sample sizes
  $n\in[50,400]$: the top-$100$ selection is stable, larger cutoffs degrade.}

  \label{fig:rock-id}
\end{figure}

\subsection{Token-Level Loss in On-Policy Distillation}
\label{sec:opd_loss}

On-policy distillation (OPD) optimizes a student policy $\pi_\theta$ by matching a teacher policy $\pi_T$ on trajectories sampled from the student's own distribution. The objective minimizes the per-token reverse KL divergence:
\begin{equation}
\mathcal{L}_{\mathrm{OPD}}(\theta) =
\mathbb{E}_{x_{1:T} \sim \pi_\theta}
\left[
\sum_{t=1}^{T}
D_{\mathrm{KL}}
\left(
\pi_\theta(\cdot \mid x_{<t})
\|
\pi_T(\cdot \mid x_{<t})
\right)
\right].
\end{equation}
For a sampled trajectory, we denote the token-level distillation signal at position $t$ as
\begin{equation}
\ell_t =
\log \pi_\theta(x_t \mid x_{<t})
-
\log \pi_T(x_t \mid x_{<t}),
\end{equation}
which measures the local student-teacher mismatch on the token actually generated by the student~\cite{fu2026revisiting}. Ideally, as the student aligns with the teacher, the magnitude of this mismatch should systematically diminish across positions and vocabulary items, leading to an optimization plateau where corrective signals eventually vanish.

However, in long-form reasoning trajectories, token-level loss does not always decrease uniformly. Even after the aggregate OPD objective saturates, certain token occurrences continue to exhibit large student-teacher divergence. Importantly, such persistent high-loss behavior may arise either from the identity of the token itself or from the local context in which the token appears. This distinction is crucial for high-frequency tokens such as whitespace, newline markers, punctuation, and numerals, which may occur many times in a single trajectory but only become high-loss in specific contexts.

\subsection{Defining and Detecting Context-Consistent Rock Tokens}
\label{sec:rock_tokens}

Contrary to the expected convergence pattern, we observe a subset of token occurrences that persistently exhibit high distillation loss even after training saturation. We refer to these as \textbf{persistent high-loss occurrences}. A token type is considered a \textbf{Rock Token} only when its persistent high-loss behavior is stable across similar contexts, rather than being caused by isolated positions in long trajectories.

We first compute the aggregate distillation burden of each vocabulary item. Since the OPD loss is accumulated over positions in student-generated trajectories, the empirical loss can be decomposed by token type $v \in \mathcal{V}$:
\begin{equation}
\mathcal{L}_{\mathrm{OPD}}(\theta)
=
\mathbb{E}_{x \sim \pi_\theta}
\left[
\sum_{t} \ell_t
\right]
=
\sum_{v \in \mathcal{V}}
\underbrace{
\mathrm{Freq}(v)
\cdot
\mathbb{E}[\ell_t \mid x_t = v]
}_{R(v)}.
\end{equation}
We define the initial \textbf{Rock Score} as
\begin{equation}
R(v) = \bar{\ell}_v \cdot \mathrm{Freq}(v),
\end{equation}
where $\bar{\ell}_v$ is the empirically estimated mean token-level loss of token $v$ at the final checkpoint, and $\mathrm{Freq}(v)$ denotes its occurrence count in the sampled trajectories. This score measures the total distillation burden associated with a token type and is used as a first-stage candidate selection criterion.

The frequency term plays an important role in reducing small-sample noise: rare tokens with extreme observed losses receive limited weight, whereas frequently occurring tokens with consistently above-baseline loss are assigned higher burden scores. Nevertheless, $R(v)$ alone is insufficient for identifying context-independent difficulty. In particular, high-frequency tokens may receive large aggregate scores even if only a small subset of their occurrences remain difficult. Therefore, we further introduce an occurrence-level and context-aware filtering procedure.

Let $o=(i,t)$ denote a token occurrence at position $t$ in trajectory $i$, with token identity $x_t^{(i)}=v$. We define the set of persistent high-loss occurrences as
\begin{equation}
\mathcal{O}_{\mathrm{PH}}
=
\left\{
(i,t):
\ell^{\mathrm{pre}}_{i,t} \ge \tau_{\mathrm{pre}},
\quad
\ell^{\mathrm{post}}_{i,t} \ge \tau_{\mathrm{post}}
\right\},
\end{equation}
where $\ell^{\mathrm{pre}}_{i,t}$ and $\ell^{\mathrm{post}}_{i,t}$ denote the token-level losses before and after OPD training, respectively. The thresholds $\tau_{\mathrm{pre}}$ and $\tau_{\mathrm{post}}$ are chosen to identify tokens that remain in the high-loss region at both stages.

To distinguish genuinely stable token-level difficulty from isolated contextual effects, we associate each occurrence with a local context window. For an occurrence $o=(i,t)$, its windowed context is defined as
\begin{equation}
c_{i,t}^{(w)}
=
x_{t-w:t+w}^{(i)},
\end{equation}
where $w$ denotes the context-window radius. We then encode this context using a context representation function $f_{\mathrm{ctx}}$:
\begin{equation}
h_{i,t}^{(w)}
=
f_{\mathrm{ctx}}
\left(
c_{i,t}^{(w)}
\right).
\end{equation}
Given two occurrences $o=(i,t)$ and $o'=(j,k)$, their contextual similarity is computed as
\begin{equation}
s(o,o')
=
\mathrm{sim}
\left(
h_{i,t}^{(w)},
h_{j,k}^{(w)}
\right),
\end{equation}
where $\mathrm{sim}(\cdot,\cdot)$ denotes a similarity function over context representations.

For each token type $v$, we compare its persistent high-loss occurrences only against other occurrences of the same token type. Let
\begin{equation}
\mathcal{O}_{\mathrm{PH}}(v)
=
\left\{
o \in \mathcal{O}_{\mathrm{PH}}:
x_o = v
\right\}.
\end{equation}
We define the context-consistency score of an occurrence $o \in \mathcal{O}_{\mathrm{PH}}(v)$ as
\begin{equation}
\rho(o)
=
\frac{1}{|\mathcal{O}_{\mathrm{PH}}(v)|-1}
\sum_{\substack{o' \in \mathcal{O}_{\mathrm{PH}}(v) \\ o' \neq o}}
\mathbf{1}
\left[
s(o,o') \ge \gamma
\right],
\end{equation}
where $\gamma$ is a contextual similarity threshold. Intuitively, $\rho(o)$ measures whether a persistent high-loss occurrence belongs to a recurring contextual pattern, rather than being an isolated spike.

We then define the set of context-consistent rock occurrences for token $v$ as
\begin{equation}
\mathcal{R}(v)
=
\left\{
o \in \mathcal{O}_{\mathrm{PH}}(v):
\rho(o) \ge \eta
\right\},
\end{equation}
where $\eta$ controls the minimum degree of contextual consistency required for an occurrence to be retained. Finally, we define the context-consistent rock rate of token $v$ as
\begin{equation}
\mathrm{CCR}(v)
=
\frac{|\mathcal{R}(v)|}{\mathrm{Freq}(v)}.
\end{equation}
This quantity measures the proportion of all occurrences of token $v$ that remain persistently high-loss and appear in similar contexts.

The final context-aware Rock Score is therefore defined as
\begin{equation}
R_{\mathrm{ctx}}(v)
=
R(v) \cdot \mathrm{CCR}(v).
\end{equation}
A token type is identified as a \textbf{Context-Consistent Rock Token} if it satisfies both a high aggregate burden and sufficient context consistency:
\begin{equation}
v \in \mathcal{V}_{\mathrm{rock}}
\quad
\Longleftrightarrow
\quad
R_{\mathrm{ctx}}(v) \ge \tau_R.
\end{equation}

This context-aware definition prevents high-frequency but context-dependent tokens from being incorrectly classified as globally stubborn. For example, whitespace, newline markers, punctuation, and numerals may occur frequently and may contribute substantially to the aggregate loss. However, if their persistent high-loss behavior appears only in a small number of isolated positions, their context-consistent rock rate $\mathrm{CCR}(v)$ will be low, reducing their final score. Conversely, tokens that repeatedly remain high-loss across similar local contexts receive high $R_{\mathrm{ctx}}(v)$ values and are retained as genuine Rock Tokens.

Operationally, our detection procedure consists of three stages. First, we rank token types by their aggregate Rock Score $R(v)$ to identify candidate high-burden tokens. Second, we identify persistent high-loss occurrences by requiring each occurrence to remain high-loss before and after OPD training. Third, we apply context-window similarity filtering to retain only those occurrences that recur in similar contexts. This produces a set of Rock Tokens whose difficulty is not merely a byproduct of frequency or isolated positional effects, but reflects stable and recurring student-teacher mismatch under comparable generation contexts.

\subsection{Empirical Analysis}
\label{sec:rock_empirical}

Rock Tokens correspond to token types whose high distillation burden is both persistent across OPD training and recurrent under similar local contexts. Following the definition in Section~\ref{sec:rock_tokens}, we first estimate the per-token KL divergence between the student policy $\pi_\theta$ and the teacher policy $\pi_T$ using student rollouts:
\begin{equation}
\widehat{\ell}_v
=
\widehat{\mathbb{E}}
\left[
D_{\mathrm{KL}}
\left(
\pi_\theta(\cdot \mid x_{<t})
\;\|\;
\pi_T(\cdot \mid x_{<t})
\right)
\mid x_t = v
\right].
\end{equation}
We then identify persistent high-loss occurrences before and after OPD training and apply context-window similarity filtering to retain only those occurrences that recur under similar local contexts. This allows us to distinguish context-consistent Rock Tokens from high-frequency tokens whose high loss is driven by isolated positions.

\textbf{Setup.}
The student model, Qwen3-4B-Instruct, is on-policy distilled from the teacher model, Qwen3-30B-A3B-Instruct, using OpenThoughts. Per-token KL statistics are sampled from OpenThoughts rollouts and evaluated on $N{=}500$ MATH-500 problems. Full experimental details are provided in Section~\ref{setup}.

\textbf{Observations.}
Figure~\ref{fig:rock-id} illustrates the distribution and density of context-consistent rock occurrences within the student's generated trajectories. Panel (a) reports the absolute count of retained rock occurrences per sequence, while Panel (b) normalizes this count by the total sequence length. After context-window similarity filtering, Rock Tokens still account for a substantial portion of the generated reasoning trajectories, with a median density of 18\% of the total tokens in each generated output.

\begin{takeaway}
\textcolor{tabblue}{\textbf{Take-away}}~
\textbf{The Pervasiveness of Context-Consistent Rock Tokens:}
Although context-aware filtering removes many high-frequency but context-dependent artifacts, the remaining Rock Tokens still occupy a substantial footprint in generation. With a median density of 18\%, nearly one-fifth of the student's reasoning trajectory consists of tokens that remain persistently high-loss and recur under similar local contexts. This suggests that student-teacher disagreement is not confined to rare outliers, but repeatedly appears in the structural scaffolding of long-form reasoning outputs.
\end{takeaway}

\subsection{Determining the Rock Token Cutoff}\label{sec:cutoff_selection}Defining the boundary of the Rock Token set $\mathcal{R}$ requires choosing a cutoff $K$ that balances \textit{selection stability} (reproducibility across samples) against \textit{loss coverage} (fraction of optimization loss captured). A small $K$ yields a stable set but low coverage, while a large $K$ captures more loss but degrades under sampling noise.

Empirical sweeps (Figures 1b and 1c) identify an optimal intersection at $K=100$. This boundary accounts for roughly 60\% of the corpus-level KL divergence while maintaining high stability across subset sizes. We therefore adopt $K=100$ to ensure a mathematically reproducible vocabulary for subsequent analyses. Comprehensive stability metrics and sweeping procedures are detailed in Appendix \ref{app:rock_ratio}.


\begin{figure*}[t]
  \centering
  \includegraphics[width=\textwidth]{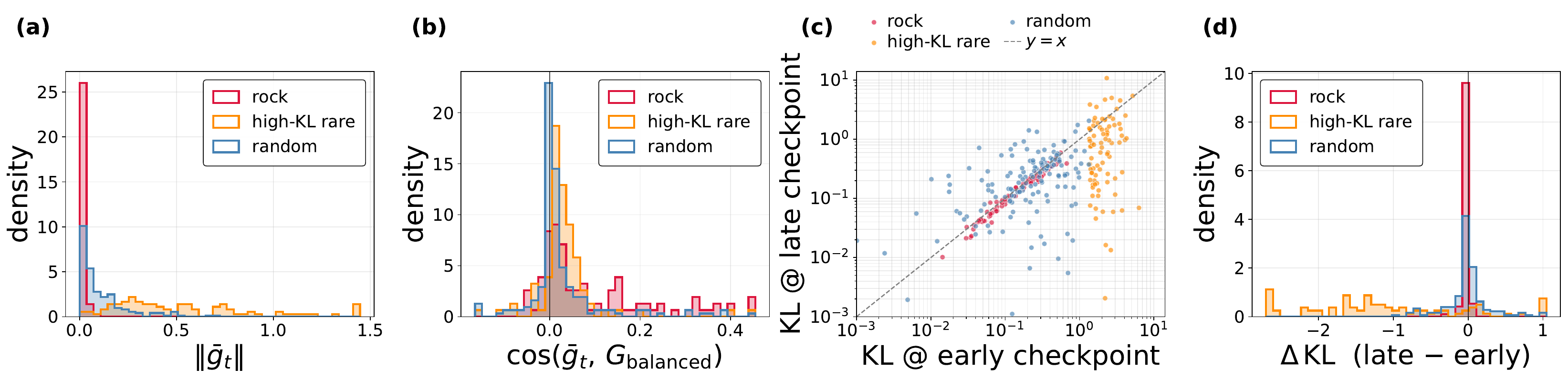}
  \caption{\textbf{Per-token gradient geometry and persistence under
  training.}
  (a) Per-token logit-gradient magnitude $\|\bar{g}_{t}\|$ by group: rocks
  are an order of magnitude smaller than rare high-KL tokens.
  (b) Cosine alignment with the frequency-balanced descent direction
  $G_{\mathrm{balanced}}$: rocks are positively aligned, with a tail
  reaching $\cos > 0.3$.
  (c) Per-token mean KL paired across two training checkpoints (log-log).
  Points below the diagonal indicate KL reduction during further training;
  rocks lie on the diagonal, rare high-KL tokens fall below.
  (d) Distribution of $\Delta\mathrm{KL} = \mathrm{KL}_{\mathrm{late}} -
  \mathrm{KL}_{\mathrm{early}}$ by group: rocks concentrate at
  $\Delta\mathrm{KL} \approx 0$ (persistent); rare high-KL tokens have
  large negative $\Delta\mathrm{KL}$ (learned).}
  \label{fig:gradient}
\end{figure*}

\subsection{Analysis of Rock Tokens}

We decoded the top-$K{=}100$ Rock Tokens into their string forms using the Qwen3 tokenizer. Despite being identified by a purely numerical criterion, the resulting set is functionally coherent: Rock Tokens consistently mark structural or discourse boundaries in mathematical reasoning. 

As detailed in Appendix~\ref{analysis_rock}, four functional clusters dominate $\mathcal{R}$: (1) \textbf{LaTeX and math delimiters}, (2) \textbf{Markdown and whitespace structure}, (3) \textbf{Discourse markers} (e.g., ``So'', ``Wait''), and (4) \textbf{Digits}. In contrast, frequency-matched controls ($\mathcal{S}_{\text{ctrl}}$) predominantly consist of content-bearing tokens (e.g., nouns, verbs) used at non-boundary positions.

\begin{takeaway}
\textcolor{tabblue}{\textbf{Take-away}}~ \textbf{Structure Over Content:} The clustering of $\mathcal{R}$ at boundaries indicates student-teacher divergence centers on \textbf{how to structure} reasoning rather than \textbf{what content} to generate. Rock Tokens are a signature of the on-policy regime, marking where the student systematically resists the teacher's discourse style.
\end{takeaway}

\subsection{Empirical Observations on Gradient Geometry and Persistence}

Having established that Rock Tokens primarily comprise syntactic and structural scaffolding, we address \textbf{RQ1: Do Rock Tokens still provide useful learning signals during optimization?} To determine whether their persistent high loss effectively translates into steady directional guidance or merely stagnant optimization noise, we characterize their training dynamics and geometric properties against rare high-KL tokens and random tokens (illustrated in Figure~\ref{fig:gradient}). 

To formally evaluate this, we analyze the geometry of per-token gradients. For each token type $t$ with $n_{t}$ occurrences, we compute the mean per-occurrence reverse-KL gradient in logit space, $\bar{g}_{t} = \tfrac{1}{n_{t}} \sum_{i} g_{i}$. We then decompose its contribution to the global descent as:
\begin{equation}
\mathrm{contrib}(t) \;=\; n_{t} \,\cdot\, \|\bar{g}_{t}\| \,\cdot\, \cos\bigl(\bar{g}_{t},\, G_{\mathrm{balanced}}\bigr),
\label{eq:contrib}
\end{equation}
where $G_{\mathrm{balanced}} = \sum_{t} \bar{g}_{t}$ is a frequency-balanced reference that grants every token type equal weight, deliberately removing the trivial frequency-weighted dominance of common tokens. This decomposition yields several key insights corresponding to our empirical observations:

\textbf{Low Gradient Magnitude vs. High Frequency (Panel a):} 
Rock tokens are characterized by substantially \emph{smaller} per-occurrence gradient magnitudes than rare high-KL tokens (median $\|\bar{g}_{t}\| \approx 0.016$ versus $\approx 0.54$; Mann-Whitney $p < 10^{-30}$). However, the multiplicative decomposition in Eq.~\ref{eq:contrib} shows that rock dominance arises from the \emph{frequency} factor ($n_t$): because of their massive occurrence rate in the training corpus, this small per-token signal accumulates into a dominant aggregate force on the weight updates.

\textbf{Alignment with Global Optimization (Panel b):} 
Counter-intuitively, the gradient direction of Rock Tokens ($\bar{g}_t$) exhibits non-trivially positive directional alignment with the balanced global gradient $G_{\mathrm{balanced}}$ (median $\cos \approx 0.040$ versus $\approx 0.025$ for high-KL and $\approx 0.006$ for random tokens), with a sub-population reaching $\cos > 0.3$. This means that each rock contributes a small but consistently aligned signal at many positions, whereas rare high-KL tokens contribute large but direction-distinct signals at very few. From the teacher's perspective, Rock Tokens do indeed point toward a ``correct'' optimization path for the overall objective.

\textbf{Persistence During Training (Panel c \& d):} 
To confirm that this directional alignment translates into resistance to further learning, we compare per-token KL between an early and late checkpoint. Rare high-KL tokens experience a substantial reduction in KL (median $2.02 \to 0.85$, $\sim 58\%$; Wilcoxon $p < 10^{-8}$), moving significantly below the $y=x$ diagonal. In contrast, Rock Tokens remain essentially unchanged (median $0.21 \to 0.19$), clustering sharply along the diagonal with the distribution of $\Delta \text{KL}$ concentrated exactly at zero (Panel d). Random control tokens show a symmetric, noisy distribution with no net change.

This optimization stagnation is geometrically consistent with our gradient findings: rare high-KL tokens possess direction-distinct gradients that focused updates can isolate and resolve. Conversely, because rocks contribute highly aligned gradients, their targeted reduction would be indistinguishable from following the global descent. Thus, they receive no separable, token-specific learning signal to overcome their inherent loss.

While our gradient decomposition explains the geometry of this stagnation, the underlying mechanical cause remains open. We hypothesize this paradox is driven by adaptive optimizers under heavy-tailed class imbalance~\cite{kunstner2024heavytailedclassimbalanceadam}. Unlike in pre-training—where Adam beneficially suppresses frequent tokens to avoid overfitting structural noise—On-Policy Distillation often requires updating these exact high-frequency structural tokens to match a teacher's formatting or style. By inflating the second-moment estimates for ubiquitous Rock Tokens, Adam systematically neutralizes their effective learning rate, trapping the student in pre-existing habits despite possessing accurate directional gradients. We leave the formal verification of this optimizer-suppression hypothesis to future work.

\begin{takeaway}
\textcolor{tabblue}{\textbf{Take-away}}~ \textbf{The Gradient Paradox:} Rock Tokens \textit{do} provide useful learning signals, generating gradients that are well-aligned with the global optimization target (Panel b). However, they remain stagnant and "untrained" throughout the OPD process (Panel d). This paradox suggests that their persistence is not due to a lack of constructive signal, but rather the student model's internal optimization bias, which stubbornly prioritizes pre-existing structural scaffolding over the teacher's stylistic direction because the rock-token signal is mathematically inseparable from the global descent direction.
\end{takeaway}

\section{Pillar Tokens: The Scaffolding of Fragile Reasoning}

\subsection{Existence as Necessity: The Functional Value of Rock Tokens}

We have observed that \textbf{Rock Tokens} remain resistant to optimization throughout the entire OPD process. To understand the origin of this resistance, we hypothesize that these tokens serve as the structural ``pillars'' of the model's reasoning process. Under this view, the student model may actively protect these tokens from optimization to prevent its internal reasoning framework from collapsing, even when they conflict with the teacher's distribution. This leads to our \textbf{RQ2: Does Rock Token persistence stem from student path dependency?}

\begin{figure}[h]
  \centering
  \includegraphics[width=\linewidth]{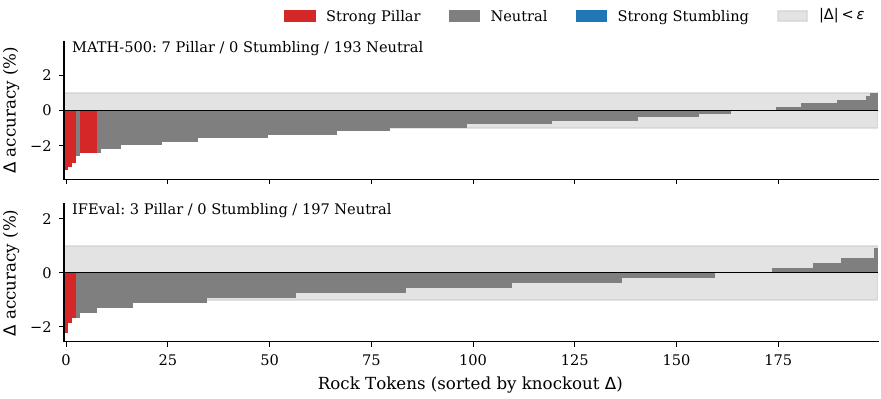}
  \caption{\textbf{Knockout effect on the $|\widetilde{\mathcal{R}}|=200$ screened Rock-Token candidates, sorted by $\Delta$.}
  Each bar is a single candidate; height is the accuracy change when its
  logit is masked at decode time. The shaded grey band marks the
  categorization threshold $|\Delta| < \varepsilon = 0.01$. Bars outside the
  band that pass the paired-bootstrap test ($\alpha=0.05$, $10{,}000$
  resamples) are coloured by category (Strong Pillar in red; Strong
  Stumbling in blue). The vast majority of candidates are Neutral on both
  benchmarks; Pillars occupy a short tail and the symmetric Strong-Stumbling
  side is empty --- the visual evidence underlying the sign-asymmetry test
  formalized below.}
  \label{fig:pillar-delta-bar}
\end{figure}
\subsection{Causal Probing via Inference-time Knockout}
\label{sec:knockout_method}

To move beyond distributional metrics and pinpoint the exact functional value of individual tokens, we employ an inference-time \textbf{knockout} procedure. We probe the causal contribution of a token $v$ by observing how its absence affects the model's accuracy on a benchmark $\mathcal{B}$. Specifically, we form a \textit{knockout policy} $\pi_\theta^{\setminus v}$ by forcing the logit of token $v$ to $-\infty$ at every decoding step. We then measure the \textbf{causal delta}:
\begin{equation}
\Delta_{\mathcal{B}}(v) = \mathrm{Acc}_{\mathcal{B}}\bigl(\pi_\theta^{\setminus v}\bigr) - \mathrm{Acc}_{\mathcal{B}}\bigl(\pi_\theta\bigr).
\label{eq:knockout_delta}
\end{equation}
A token $v$ is identified as a \textbf{Strong Pillar} if its removal causes a significant degradation ($\Delta_{\mathcal{B}}(v) \le -0.01$). We screen a pool of $|\widetilde{\mathcal{R}}|=200$ candidates, extending beyond the core $K=100$ set to ensure a wider search for these rare causal effects (see Appendix~\ref{sec:pillar-experiments} for statistical and operational details).

\subsection{Are Rock Tokens Pillars? Findings and Orthogonality}
\label{sec:pillar_findings}

The distribution of $\Delta_{\mathcal{B}}(v)$ in Figure~\ref{fig:pillar-delta-bar} reveals the underlying nature of the Rock Token set. Our investigation yields three primary observations:

\textbf{Pillars are rare but vital:} Only $3.5\%$ of Rock Tokens on MATH-500 and $1.5\%$ on IFEval qualify as Strong Pillars. While most high-KL tokens are neutral residuals, a small subset is absolutely indispensable for the model's reasoning trajectory.
\textbf{Directional Asymmetry:} The complete absence of Strong Stumbling Blocks suggests that the student's "stubborn" deviations from the teacher are rarely detrimental to its own logic; they are either necessary anchors or harmless stylistic preferences.
\textbf{Orthogonality to Traditional Metrics:} 
To examine if Pillarhood can be predicted by standard signals, we analyzed the correlation between $\Delta_{\mathcal{B}}(v)$ and several distributional metrics, including student/teacher entropy, log-frequency, and residual KL. 
Crucially, none of these predictors exhibit a significant correlation with a token's causal importance ($|r| < 0.07$, see Appendix~\ref{sec:pillar-vs-forking} for full analysis and scatter plots). 
This demonstrates that Pillarhood is a property orthogonal to existing importance taxonomies~\cite{tip2026}, suggesting that importance schemes keyed to loss or entropy signals risk suppressing the very tokens the student model cannot reason without.

\begin{takeaway}
\textcolor{tabblue}{\textbf{Take-away}}~
\textbf{The Concentration of Necessity.} 
While Rock Tokens as a population are indispensable for training stability, their causal necessity at inference is highly concentrated: only a tiny fraction ($1.5\%$--$3.5\%$) act as true \textbf{Pillars}. 
This refutes the hypothesis that the student relies on the \textit{entire} high-loss set to maintain reasoning; instead, the model appears to protect a wide "buffer zone" of persistent residuals to ensure a few critical, semantic anchors remain intact.
\end{takeaway}

\section{The genuine contribution of Rock Tokens} 
\label{sec:rq3}

%

\begin{wrapfigure}[11]{r}{0.47\columnwidth}
    \vspace{-1.2em}
    \centering
    \includegraphics[width=0.45\columnwidth]{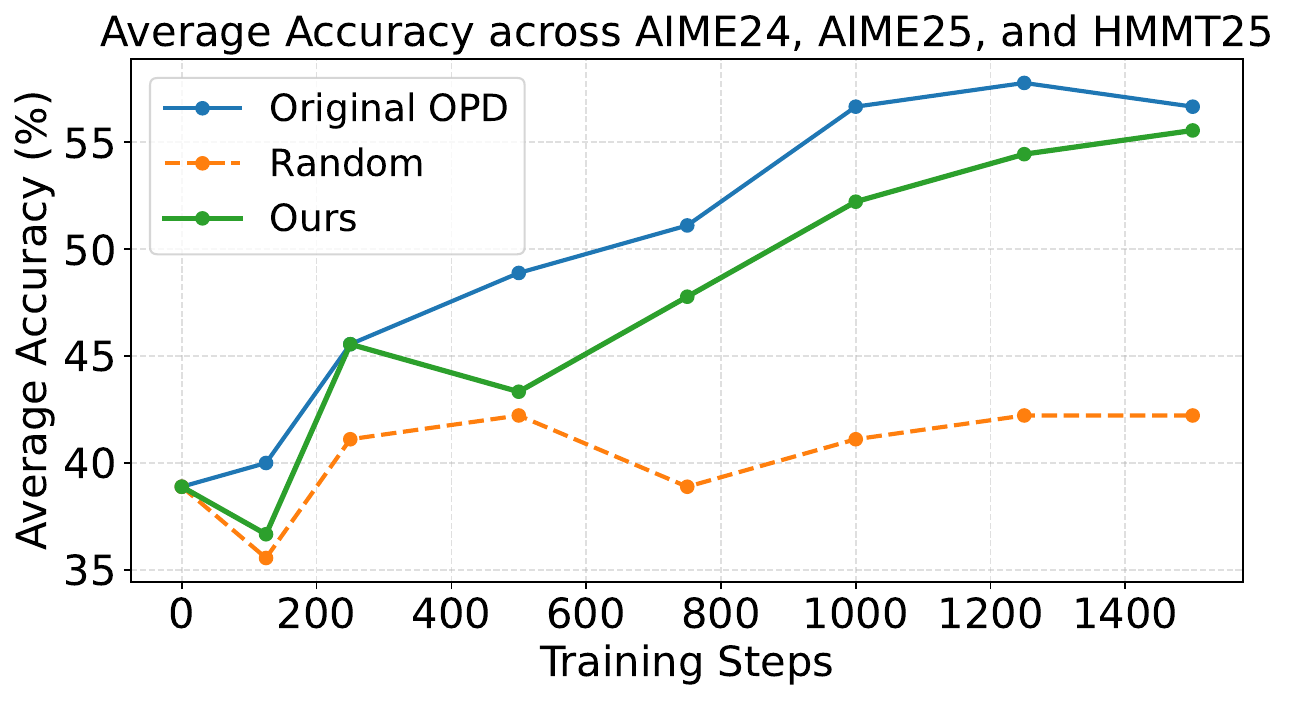}
    \vspace{-0.8em}
    \caption{
    Average accuracy across AIME24, AIME25, and HMMT25 during OPD training. Each 200 training steps correspond to 8,000 prompts, with 4 rollouts per prompt.
    }
    \label{fig:opd_random_vs_ours}
    \vspace{-1.2em}
\end{wrapfigure}
\leavevmode\par
\noindent

\subsection{RQ3: What is the genuine functional contribution of Rock Tokens to model training?}

The functional redundancy of Rock Tokens at inference prompts a critical question: do their persistent high-loss signals provide essential constraints, or are they merely \textbf{optimization noise}? We consider two competing hypotheses: (1) \textbf{Late-stage Saturation}, where these tokens act as early alignment drivers but become unproductive "Stumbling Blocks" as the model reaches structural limits; or (2) \textbf{Inherent Inutility}, where they represent superficial stylistic gaps with no reasoning value from the outset. 

To adjudicate between these views, we employ a \textbf{selective distillation strategy} by freezing the gradients of Rock Tokens. By comparing training trajectories with and without these signals, we aim to determine whether these persistent residuals are indispensable functional anchors or a detrimental optimization burden that consumes gradient bandwidth without yielding performance gains.

\subsection{Selective Distillation via Window-Aware Token Reweighting}
\label{Preliminaries:Stumbling Tokens and Token-Level Reweighting}

To evaluate the functional necessity of these persistent residuals, we introduce a window-aware token reweighting mechanism to the OPD objective. Since Rock Tokens are identified through local high-divergence windows rather than isolated token-level spikes, we reweight not only individual candidate tokens but also the local trajectory segments in which they appear. Let $\mathcal{R}$ denote the Persistence Set defined in Section~\ref{sec:rock_tokens}, and let $W(v)$ denote the local high-divergence window associated with each candidate token $v\in\mathcal{R}$. We define the union of Rock-associated windows as
\begin{equation}
\mathcal{W}_{\mathcal{R}} = \bigcup_{v\in\mathcal{R}} W(v).
\end{equation}
The weighted OPD objective is then given by
\begin{equation}
\mathcal{L}_{\text{weighted}}
=
\mathbb{E}_{x \sim \pi_\theta}
\sum_{t=1}^{T}
w(x_t, t) \cdot \ell_t,
\quad \text{where }
w(x_t,t)
=
\begin{cases}
\lambda, & x_t \in \mathcal{R} \ \text{or} \ t \in \mathcal{W}_{\mathcal{R}}, \\
1, & \text{otherwise}.
\end{cases}
\label{eq:window-aware-reweighting}
\end{equation}
Here, $\lambda \in [0,1]$ serves as a scaling factor to modulate the optimization pressure assigned to Rock-associated tokens and their local high-divergence contexts. This formulation allows us to test whether the gradient bandwidth consumed by high-Rock-score regions is productive for learning or merely an optimization burden caused by persistent teacher--student stylistic or trajectory-level mismatch.

Our core experimental design employs three regimes by manipulating $\lambda$. The \textbf{Baseline OPD} ($\lambda=1$) follows standard distillation without intervention. \textbf{Rock-Freeze (Our Method)} sets $\lambda=0$ for all tokens in $\mathcal{R}$ and their associated high-divergence windows $\mathcal{W}_{\mathcal{R}}$. This ``Just Not Train'' strategy masks the gradients of Rock-associated local regions, allowing us to determine whether these residuals function as indispensable reasoning anchors or as unproductive stumbling regions that can be bypassed to save optimization budget. Finally, the \textbf{Freq-Matched Window Freeze (Random)} applies the same $\lambda=0$ strategy to a randomly sampled set of windows $\mathcal{W}_{\mathcal{S}}$ whose token frequency distribution and window-length distribution are matched to $\mathcal{W}_{\mathcal{R}}$. By comparing these three settings, we can decouple the effect of structurally persistent Rock-associated regions from stochastic noise, frequency-related artifacts, and window-size effects.

This window-aware reweighting also connects the intervention to our Pillar/Stumbling distinction. If freezing Rock-associated windows improves or preserves downstream performance, the corresponding regions are more likely to behave as \textbf{Stumbling Blocks}: high-divergence segments that consume gradient budget without supporting successful reasoning. Conversely, if freezing them substantially degrades performance, the regions are more consistent with \textbf{Pillar Tokens} or pillar-like local segments, suggesting that their high divergence reflects a functional reasoning pattern that should not be suppressed.

\subsection{Impact of Rock Tokens on Reasoning Performance}

Figure~\ref{fig:opd_random_vs_ours} illustrates the performance trajectories across AIME24, AIME25, and HMMT25 under different intervention regimes. Our analysis reveals three key findings. First, while Rock Tokens exhibit persistent high loss, they are not merely "noise traps"; their inclusion provides essential gradients that anchor the optimization process, as evidenced by the performance gap between the \textit{Original OPD} and \textit{Ours}. Second, while reducing total gradient signals is generally detrimental—shown by the severe degradation in the \textit{Random} baseline—selectively down-weighting Rock Tokens is significantly safer, maintaining a much higher accuracy ceiling. Third, in our setup, Rock Tokens account for approximately 18\% of the total output tokens; by mitigating the optimization pressure on these residuals, our empirical results demonstrate a 1.7$\times$ increase in overall training wall-clock speed without substantial loss in reasoning integrity.

\begin{takeaway}
\textcolor{tabblue}{\textbf{Take-away}}~
\textbf{The Efficiency-Performance Frontier:} While Rock Tokens provide necessary structural gradients, they exhibit significantly diminishing marginal utility. Our results demonstrate that applying a freeze-weighting to these 30\% high-cost tokens yields a \textbf{1.4$\times$ wall-clock speedup} with minimal impact on reasoning integrity. This confirms that while Rock Tokens are not entirely disposable, \textbf{non-uniform gradient allocation} can effectively bypass computational redundancy in on-policy alignment.
\end{takeaway}

\section{Experiment Setups}
\label{setup}
\subsection{Off-policy and On-policy Distillation}
\label{sec:training_setup}

We implement a two-stage distillation pipeline via the \textbf{KDFlow} framework~\cite{zhang2026kdflow} to transfer knowledge from a teacher, \textbf{Qwen3-30B-A3B-Instruct-2507} (MoE, 3B active), to a student, \textbf{Qwen3-4B-Instruct-2507}~\cite{qwen3technicalreport}, with thinking mode disabled for both. 
\textbf{Stage 1 (Off-Policy):} To achieve initial \textit{all-vocabulary alignment}, the student is trained for one epoch on 20k teacher-generated solutions from \textbf{OpenThoughts3}~\cite{guha2025openthoughts} using a forward KL objective ($\text{kd\_ratio}=0.5$) with cross-entropy. 
\textbf{Stage 2 (On-Policy):} Starting from the Stage 1 checkpoint, the student samples 4 rollouts each prompt from a separate 10k prompts as a checkpoint, (7 checkpoints in total)  and is trained for one epoch on its own trajectories using pure reverse KL ($\text{kd\_ratio}=1.0$). 
Unless specified, all other hyperparameters follow KDFlow defaults; see Appendix~\ref{app:training_details} for full implementation details.
\subsection{Evaluation}
We employ the \textsc{lm-evaluation-harness} framework \cite{eval-harness} for standardized assessment across all benchmarks in a zero-shot setting. To ensure statistical robustness, we report the \textbf{Pass@1} accuracy averaged over five independent runs, accounting for variance in decoding.

Our evaluation suite focuses on challenging competitive mathematics, comprising \textbf{AIME 24}, \textbf{AIME 25}, and \textbf{HMMT 25-Feb} from MathArena~\cite{dekoninck2026matharena}. Each benchmark contains 30 problems; we report the aggregated average score across these 90 tasks as the primary performance metric in our study. We extend our analysis to MATH-500~\cite{lightman2023letsverifystepstep} and IFEval~\cite{zhou2023instruction} to obtain a robust sample size for token-level statistical evaluation.

\section{Related Works}

\subsection{On-Policy Distillation for LLM Post-Training}

Among the various capabilities of LLMs~\cite{li-etal-2025-generation,zhang2025find,li2025preference}, reasoning has emerged as a central focus~\cite{xu2025learning,jiang2026scribestructuredmidlevelsupervision,tong2024can,Li2025Efficient}. Traditional methods for improving reasoning include supervised SFT~\cite{li2024contextualization,jiang2025drp,li2025ddtime,zhou2024adversarial} and RL~\cite{guo2025deepseek,xu2025learning}.

On-policy distillation evolves beyond SFT and RLVR by combining trajectory exploration with dense, token-level guidance~\cite{lu2025onpolicydistillation}. Recent large-scale systems like DeepSeek-V4~\cite{deepseekv4}, MiMo~\cite{xiao2026mimo}, and Qwen-3~\cite{qwen3technicalreport} demonstrate its efficacy as a post-training stage, while self-distillation studies~\cite{zhao2026self,shenfeld2026self,hubotter2026reinforcement} suggest OPD is a general mechanism for amplifying latent model capabilities rather than mere imitation. However, the success of OPD is often hindered by teacher-student misalignment. Prior research examines this incompatibility through global factors such as vocabulary, reasoning style, teacher selection, and model family~\cite{fu2026revisiting,li2026rethinking,zhang2025find}. 

Recent work also connects distillation with broader efficiency-oriented post-training objectives. Existing studies explore reasoning pruning, long-context compression, layer partitioning, data-efficient compression, and model-scale efficiency trade-offs~\cite{jiang2025drp,11460600,li2024sglp,10800533,cao2026taskspecificefficiencyanalysissmall,zhang2026performance}. Related distillation and alignment strategies have further been extended to multimodal, temporal, and structured representation settings~\cite{li2025frequency,11460474,li2025ammkd}. 

Our work focuses on a fine-grained source of stagnation in OPD. We investigate \textbf{Rock Tokens}---persistent high-loss positions that resist alignment even after training plateaus---to determine when OPD supervision provides productive signals versus unproductive noise.

\subsection{Token-Level Learning Dynamics in Reasoning Models}

Recent studies demonstrate that reasoning in LLMs is disproportionately supported by critical ``anchor'' tokens, the perturbation of which during inference significantly degrades model performance~\cite{gupta2025llms,li2025mmt,xu2026stable}. In parallel, RLVR research reveals that training is non-uniform~\cite{yang2025token,cheng2026reasoning,wu2025mitigating,al2026dagger}, concentrating optimization on high-entropy ``forking tokens'' that act as decisive branching points for reasoning gains~\cite{wangbeyond}, and further analysis confirms the overall token distribution drift mainly happens on sparse but crucial tokens~\cite{meng2026sparse,luo2025cellink}. 

This token-level perspective is also related to recent work on structured reasoning, tool-use supervision, and evaluation. Studies on tool-integrated reasoning and structured mid-level supervision show that reasoning improvements often depend on how intermediate steps are organized and supervised~\cite{xu2025learning,jiang2026scribe,li2025srkd}. Work on LLM-as-a-judge, RAG faithfulness, and evidence calibration similarly suggests that local signals can become unreliable when detached from broader reasoning or contextual structure~\cite{li2025generation,chen2026doesragknowretrieval,qian2026relevantwarrantedevidenceforcecalibration,cheng2026resolvingrobustnessprecisiontradeofffinancial,zhou2022understanding}. 

Beyond text-only reasoning, related alignment problems also appear in multimodal retrieval and structured representation learning. Prior work studies fine-grained semantic grounding, compositional matching, and robust alignment under complex modification signals~\cite{FineCIR,ENCODER,HABIT}. Other studies investigate anchor-based calibration, noise-unlearning, and attribute-aware retrieval representations~\cite{TEMA,ConeSep,li2024comae}, while video retrieval and arbiter-calibrated retrieval further highlight the importance of stable alignment under changing evidence or modality conditions~\cite{ReTrack,Air-Know,miao2026seeing}. These studies are not direct OPD methods, but they reflect a broader trend: effective learning often requires identifying which local signals are structurally meaningful rather than treating all supervision signals uniformly.

Although OPD's dense supervision suggests high-loss tokens drive training, their specific contribution mechanism remains unclear---especially when they persist as \textbf{Rock Tokens}. It is unknown whether these recalcitrant positions are indispensable reasoning pillars or merely unproductive residues. We address this gap by systematically analyzing their gradient dynamics and training effects to determine their true role in the OPD process.

\subsection{Broader Efficient and Structured Learning Systems}

Recent developments in agentic and structured AI systems further motivate examining when local supervision signals help or hinder learning. Work on agentic memory and multi-agent systems studies how memory structure, recall, reputation, and coordination affect reasoning reliability~\cite{jiang2026magma,jiang2026anatomy,liu2026memory,10.1007/978-981-95-4384-7_10,zhou2024boosting}. Related research on quantized multimodal models, medical image segmentation, speech separation, facial expression recognition, and 3D interaction systems shows that efficiency-oriented learning problems often require balancing compression, robustness, and task-specific structure~\cite{guo2025quantized,li2026sepprune,10.1117/12.3096291,li2026comprehensive}.  Efficient retrieval, reasoning, and representation learning in domain-specific settings. Existing work investigates robustness-precision trade-offs and reranking strategies in financial RAG systems~\cite{cheng2026resolvingrobustnessprecisiontradeofffinancial,cheng2026enhancingfinancialreportquestionanswering}, energy-efficient RAG architectures for small language models~\cite{cheng2026toward,xie2025chat}, and semantic embedding analysis for short-text understanding~\cite{11484350,xie2026delving}. Related applications further include LLM-based financial disclosure analysis~\cite{liu2026improving,xie2026hvd}, co-design frameworks for efficient multimodal inference~\cite{chen2025autoneuralcodesigningvisionlanguagemodels}, and time-series studies on volatility forecasting and regime-dependent market dynamics~\cite{cheng2026volatility,Cheng2026}.
Structured data and domain-specific applications provide additional evidence that model behavior is often shaped by sparse but influential signals. Recent studies explore graph-enhanced spreadsheet understanding, graph search, nearest-neighbor retrieval, spatiotemporal prediction, financial sentiment forecasting, and adaptive robustness under distribution shifts~\cite{lei2026sheet,wang2023towards,WangYZLCL24,zeng2025enhancing,zhang2026finsentllm,wu2026adaptive}. In applied financial and business domains, LLM-based or representation-learning methods have also been used for disclosure analysis, market behavior modeling, greenwashing analysis, and short-text semantic understanding~\cite{liu2026improving,dai2023neighbors,dai2023analyst,11484350}. 

These broader studies reinforce the motivation for our analysis: rather than assuming all difficult tokens are equally useful, it is important to distinguish productive learning signals from persistent, misaligned, or noisy supervision. Our work contributes to this direction by isolating and characterizing Rock Tokens within OPD training dynamics.

\section{Conclusion}
This paper investigates the role of high-loss tokens in On-Policy Distillation, revealing that while they occupy a dominant position in the loss landscape, they are not the decisive factor for performance gains. Our analysis identifies a distinct category of persistent tokens, termed \textbf{Rock Tokens}, despite their significant gradient magnitude and apparent "effort" during optimization, Rock Tokens fail to provide constructive guidance for the model's reasoning capabilities. Notably, our intervention experiments demonstrate that applying a freeze-weighting to these tokens—thereby deprioritizing them during the optimization update—results in negligible impact on downstream performance while achieving a 1.4$\times$ increase in training wall-clock speed. These findings challenge the conventional wisdom that aligning every high-loss token is indispensable for learning. Instead, we advocate for a more resource-efficient, token-selective approach that bypasses the computational redundancy of low-utility tokens to optimize the efficiency-performance frontier in on-policy alignment.

{
\small
\bibliographystyle{abbrvnat} 
\bibliography{ref_raw}
}

\medskip


\appendix

\section{Limitations}
First, our empirical evaluations rely predominantly on competitive mathematical reasoning; the functional role of Rock Tokens may differ in open-ended generation or coding tasks where structural boundaries are less rigid. Second, our experimental setup utilizes a specific 30B-to-4B distillation pairing, leaving the persistence and geometry of these tokens across different architectural families or vastly larger scales an open question. Finally, to isolate the functional impact of Stumbling Blocks, our mitigation strategy employs a strict binary gradient-freeze mask. While effective for demonstrating causal necessity, this approach is a blunt instrument; future work should explore dynamic reweighting schemes or soft-penalties that gracefully adapt to token-level loss trajectories throughout the optimization process.
\subsection{Ethics Discussion and Risk Assessment}

This work studies model distillation and training dynamics, and does not introduce a deployed system, human-subject data, or sensitive personal data. Its direct societal risks are therefore limited. However, because improvements in language model training may be used in both beneficial and harmful downstream applications, we acknowledge potential dual-use risks, including misuse for generating low-quality or misleading content.

We mitigate these risks by limiting the paper to research analysis and experimental evaluation. Any released code, models, or processed artifacts will be accompanied by appropriate documentation, intended-use information, and license details. All pretrained models, datasets, and software frameworks used in this work are credited, and their licenses and terms of use are respected.

\section{Analysis on Rock Tokens}
\label{analysis_rock}
 We decode the top-$K{=}100$ rock tokens and their frequency-matched controls $\mathcal{S}_{\text{ctrl}}$ into their string forms using the public Qwen3 tokenizer. Despite being identified by a purely numerical criterion ($R(v)$), the resulting set is functionally coherent: rock tokens fall into a small number of identifiable categories, all of which mark structural or discourse boundaries in mathematical reasoning output.

Four functional clusters dominate $\mathcal{R}$.
\begin{enumerate}
\item \textbf{LaTeX and math delimiters} — \$', \verb|\\|, \verb|$$|, \verb|=|, \verb|^|, \verb|{|, \verb|}|, \verb|frac|, \verb|(|, \verb|)|. These mark transitions into and out of math mode and operate at the syntactic boundary between prose and formula.   \item \textbf{Markdown and whitespace structure} — **', \textbackslash n', \textbackslash n\textbackslash n', :\textbackslash n\textbackslash n', ---\textbackslash n\textbackslash n', \#\#\#'. These delimit logical units (paragraphs, section headers, bold spans).   \item \textbf{Discourse markers at sentence-initial position} — So', Let', We', But', Now', Wait', Then', Since', This'. These open new reasoning steps and signal turns in the chain of thought.
\item \textbf{Digits} — every digit 0' through 9' appears in $\mathcal{R}$, each with frequency in the thousands and a non-trivial mean per-token KL. These are tokens that punctuate quantitative statements.
\end{enumerate}

Contrast with the frequency-matched control set. The control set $\mathcal{S}_{\text{ctrl}}$, sampled at the same frequency distribution, is qualitatively different: it consists predominantly of content-bearing tokens used at non-boundary positions — common nouns and verbs of mathematical discourse (e.g.\ triangle', equation', region', expression', must', find', check'), prepositions and conjunctions used mid-clause (as', between', first', right'), and single-letter variable tokens (a', b', i', f', y'). The frequency-matching procedure thus isolates the relevant axis: rocks and controls share the quantity of supervision they receive during OPD, but differ sharply in their syntactic role.

\textbf{Interpretation} The clustering of $\mathcal{R}$ at structural boundaries indicates that the residual student–teacher divergence after on-policy distillation is concentrated at the points where the model decides \emph{how to structure} its reasoning output, rather than \emph{what content} to emit. Decisions like open math mode'', begin a new paragraph'', insert a section header'', or start a new reasoning step with So' versus Then','' are precisely the positions where the OPD loss persists, even after extensive distillation. This observation suggests that rock tokens are not arbitrary statistical outliers but a consistent signature of the on-policy distillation regime: they identify the structural/discourse decisions that the student systematically fails to align with the teacher.
\section{Training and Evaluation Details}
\label{app:training_details}
\subsection{off policy and on policy training}

\paragraph{Training Details.} We distill from Qwen3-30B-A3B-Instruct-2507, a Mixture-of-Experts teacher with $\sim$30B total and $\sim$3B active parameters, into Qwen3-4B-Instruct-2507 as the student, using the KDFlow framework. Both models are run with thinking mode disabled (\texttt{enable\_thinking=False}). Training proceeds in two stages on a single node of $4\times$ H100 (80\,GB) GPUs with FSDP2, bf16, and gradient checkpointing.
  
\paragraph{Stage 1 -- Off-policy KD.} We first sample 20k teacher responses on math prompts drawn from OpenThoughts3 ($\text{temperature}=0.6$, $\text{top\_p}=0.95$, $\text{max\_new\_tokens}=16384$, $\text{TP}=2$ on $2\times$ H100). The student is then trained on these (prompt, teacher-response) pairs with a per-token KL distillation loss combined with cross-entropy at $\text{kd\_ratio}=0.5$ (vanilla KD, forward KL). We use AdamW with learning rate $2\times10^{-5}$, $5\%$ linear warmup, global batch size $128$ (micro-batch $1$), max sequence length $16384$, sample packing, and ring attention of size $2$, for $1$ epoch.

\paragraph{Stage 2 -- On-policy KD.} Initialized from the Stage-1 checkpoint, the student generates $4$ rollouts per prompt ($\text{temperature}=1.0$, $\text{top\_p}=1.0$, $\text{generate\_max\_len}=8000$, $\text{prompt\_max\_len}=800$) on a separate 10k-prompt slice (positions 20k--30k of the same source), and is distilled toward the teacher's token distributions on those rollouts. We use vanilla KD with reverse KL at $\text{kd\_ratio}=1.0$ (pure distillation, no CE), learning rate $2\times10^{-6}$, $5\%$ linear warmup, gradient clipping at $1.0$, global batch size $4$ (micro-batch $1$), and $1$ epoch. The rollout engine uses $1$ engine with $\text{TP}=2$ and the teacher uses $\text{TP}=4$; both engines share GPUs with the trainer via offload-to-CPU sleep/wakeup (\texttt{teacher\_enable\_sleep=True}, \texttt{rollout\_enable\_sleep=True}.

\paragraph{Defaults.} All other hyperparameters are KDFlow defaults; we did not tune optimizer betas, weight decay, dropout, or KD temperature (kept at $1.0$).

\section{Pillar Correlation and Multiple-Testing Analysis}
\subsection{Pillar Correlation Analysis}
\label{app:pillar_correlations}

Table~\ref{tab:pillar-correlations} reports the full set of Pearson correlations between the inference-time knockout effect $\Delta_{\mathcal{B}}(v)$  and nine candidate token-level predictors, measured over the $|\widetilde{\mathcal{R}}|=200$ screened Rock-Token candidates of \S\ref{sec:pillar-experiments} and reported separately for MATH-500 and IFEval. The figure in the main text (Figure~\ref{fig:pillar-vs-predictors}) visualizes a representative subset of six of these predictors on MATH-500; this table is the complete record, including IFEval and the three loss-based features (pre-/post-OPD KL and KL improvement). Across all eighteen $(predictor, benchmark)$ pairs, no correlation reaches $|r| > 0.075$ and no $p$-value falls below $0.30$. The largest observed magnitude on either benchmark is the IFEval correlation with KL improvement ($r{=}+0.071$, $p{=}0.32$), which is well within the noise band expected at $n{=}200$. We therefore cannot reject the null hypothesis $r=0$ for any predictor on either benchmark, supporting the claim in \S\ref{sec:pillar-vs-forking} that Pillarhood is statistically orthogonal to the entropy-, frequency-, and divergence-based importance signals used by prior work~\cite{wangbeyond,tip2026,yang2025token,wu2025mitigating}.

\begin{table}[h]
\centering
\caption{Pearson correlation between the per-token knockout effect $\Delta_{\mathcal{B}}(v)$ and nine candidate predictors over the $|\widetilde{\mathcal{R}}|=200$ screened Rock-Token candidates, reported for MATH-500 and IFEval. $p$-values are two-sided. No correlation reaches $|r|>0.075$ on either benchmark; no $p$-value falls below $0.30$.}
\label{tab:pillar-correlations}
\small
\begin{tabular}{lcccc}
\toprule
\textbf{Predictor} & \multicolumn{2}{c}{\textbf{MATH-500}} & \multicolumn{2}{c}{\textbf{IFEval}} \\
\cmidrule(lr){2-3}\cmidrule(lr){4-5}
 & $r$ & $p$ & $r$ & $p$ \\
\midrule
Frequency                  & $+0.015$ & $0.832$ & $+0.030$ & $0.677$ \\
Rock count                 & $+0.013$ & $0.855$ & $+0.023$ & $0.751$ \\
Rock rate                  & $-0.022$ & $0.760$ & $-0.047$ & $0.512$ \\
Pre-OPD KL                 & $-0.017$ & $0.808$ & $-0.015$ & $0.833$ \\
Post-OPD KL                & $+0.010$ & $0.883$ & $-0.048$ & $0.503$ \\
KL improvement             & $-0.062$ & $0.386$ & $+0.071$ & $0.320$ \\
Teacher entropy            & $+0.063$ & $0.378$ & $-0.036$ & $0.611$ \\
Pre-OPD student entropy    & $+0.059$ & $0.406$ & $-0.069$ & $0.331$ \\
Post-OPD student entropy   & $+0.056$ & $0.432$ & $-0.037$ & $0.604$ \\
\bottomrule
\end{tabular}
\end{table}

\subsection{Multiple-Testing Analysis for the Per-Token Pillar Criterion}
\label{app:pillar_multiple_testing}

The per-token Pillar criterion of \S\ref{sec:pillar-identification} applies a paired bootstrap test at $\alpha=0.05$ independently to each of the $|\widetilde{\mathcal{R}}|=200$ screened Rock-Token candidates, conjoined with the effect-size filter $|\Delta_\mathcal{B}(v)|\ge\varepsilon$ ($\varepsilon=0.01$). Because $\varepsilon$ is approximately at the per-token sampling-error scale, the conjunction does not appreciably reduce the per-test Type-I rate, and $200\!\times\!0.05 = 10$ rejections are expected under the global null. We report below how the per-token identifications behave under standard family-wise and false-discovery corrections, and the global sign-asymmetry test on which the population-level claim of \S\ref{sec:pillar-experiments} rests.

\paragraph{Per-token corrections.} Bonferroni control of the family-wise error at $\alpha=0.05$ requires $p<\alpha/200=2.5\times 10^{-4}$. The smallest observed $p$-values are $0.0038$ (MATH-500, ``\,certain'') and $0.0016$ (IFEval, ``-like''); neither passes Bonferroni, and consequently no token does. Benjamini--Hochberg control of the false discovery rate at $q\in\{0.05, 0.10, 0.20\}$ likewise produces zero rejections on either benchmark: the BH cutoff $p_{(k)}\le (k/n)\,q$ is not crossed even for the smallest-$p$ token, because $1/200 \times 0.20 = 10^{-3} < 0.0016$. The per-token Pillar identifications are therefore \emph{exploratory candidates}, not confirmatory rejections in the multiple-testing sense.

\paragraph{Global sign-asymmetry test.} The exploratory candidates do, however, exhibit a non-trivial structural property: the sign of the significant rejection is consistent across benchmarks. Of the seven MATH-500 candidates, all seven satisfy $\Delta_{\mathrm{MATH500}}(v)<0$ (Pillar direction); of the three IFEval candidates, all three satisfy $\Delta_{\mathrm{IFEval}}(v)<0$. Combined, the rejection sign-distribution is $10$ negative, $0$ positive. Under the global null in which every per-token rejection is a chance event, the rejection sign should be Bernoulli$(\nicefrac{1}{2})$, and the probability of observing all $10$ on one side is $2^{-10}\approx 9.8\times 10^{-4}$ (two-sided exact binomial $p = 1.95\times 10^{-3}$). On MATH-500 alone the analogous sign test is $7/0$ with two-sided $p=0.016$; on IFEval alone $3/0$ with $p=0.25$. The combined test is a single global test that does not inflate with the number of tokens screened, and it is the result on which \S\ref{sec:pillar-experiments} relies for the population-level claim that a non-empty subset of Rock Tokens is causally necessary at decode time.

\paragraph{Summary.} The reader is advised to read the per-token Pillar identifications of \S\ref{sec:pillar-experiments} (e.g.\ ``\,certain'', ``\,strategic'', ``\,Initialize'') as the strongest individual candidates from a population-level effect, rather than as $7$ independently-confirmed discoveries. The empirical contribution of this section that survives strict multiple-testing scrutiny is the sign asymmetry, which is incompatible with the multiplicity-noise null at $p\approx 0.002$.

\paragraph{Stable-core robustness check.} The screening pool of \S\ref{sec:pillar-identification} is the top-$K_{\text{screen}}=200$ count-ranked extension of $\mathcal{R}$, while the definitional Rock-Token set fixed by the stability/coverage analysis of \S\ref{ref:recalcitrant_tokens} is the top-$K=100$. The 10 Strong Pillars from the bootstrap distribute as follows over the count-rank index. On MATH-500 (ranks reported on the top-200 list): ``\,certain'' (16), ``\,Initialize'' (93), ``\,strategic'' (99), ``Do'' (109), ``\,programs'' (143), ``\,tech'' (166), ``\,arms'' (185). On IFEval: ``-like'' (35), ``\,tools'' (64), ``\,trade'' (69). Six of the ten lie within the $K=100$ core, including the smallest-$p$ token on each benchmark (``\,certain'' at $p=0.0038$ on MATH-500 and ``-like'' at $p=0.0016$ on IFEval). The remaining four lie in the rank-$109$-to-$185$ regime, which \S\ref{ref:recalcitrant_tokens} flags as below the Jaccard-stability sweet spot; we therefore caution that those four are at greater risk of being corpus-specific. The sign-asymmetry test of the previous paragraph is robust to this concern: restricted to the $K=100$ stable core, it remains $6/0$ negative-versus-positive, with two-sided exact-binomial $p = 2^{-6}\!\times\!2 = 3.1\times 10^{-2}$.

\subsection{Definition and Identification of `Pillar Tokens'}
\label{sec:pillar-identification}

To investigate this, we operationally categorize Rock Tokens based on their causal impact at inference time. Since Rock Tokens are identified through a sliding-window criterion rather than isolated token-level loss spikes, we define Pillarhood in a window-aware manner. Specifically, each candidate token $v\in\widetilde{\mathcal{R}}$ is associated with a local high-divergence window $W(v)$, which denotes the surrounding decoding context in which the teacher--student divergence remains consistently elevated. \textbf{Pillar Tokens} are defined as tokens whose removal or perturbation, either individually or together with their local high-divergence window, degrades task accuracy, suggesting that the model \emph{cannot reason without the local trajectory segment they support}. In contrast, candidates whose removal improves accuracy are categorized as \textbf{Stumbling Blocks}, while the remainder are deemed functionally neutral.

\emph{\textbf{Intuition.}} The core idea is that if a token is truly load-bearing for reasoning, deleting it from the decoder should break the trajectory; if it only \emph{looks} ``hard'' due to high loss but lacks functional utility, its deletion should change nothing. Importantly, because high-loss tokens often appear as part of a short locally unstable reasoning segment, we do not interpret Pillarhood solely from an isolated token-level spike. Instead, we use a window-aware knockout test: we first perturb the candidate token itself, and then perturb its surrounding high-divergence window $W(v)$ to determine whether the local segment is causally necessary for successful reasoning.

Given the post-OPD student $\pi_\theta$ and a candidate $v\in\widetilde{\mathcal{R}}$ (screening pool defined below), we first form a token-level knockout policy $\pi_\theta^{\setminus v}$ that sets the logit of $v$ to $-\infty$ at every decoding step. We then record the token-level accuracy delta on a held-out benchmark $\mathcal{B}$:
\begin{equation}
\Delta_{\mathcal{B}}(v) =
\mathrm{Acc}_{\mathcal{B}}\bigl(\pi_\theta^{\setminus v}\bigr)
-
\mathrm{Acc}_{\mathcal{B}}\bigl(\pi_\theta\bigr).
\label{eq:pillar-delta-token}
\end{equation}

To account for the fact that Rock Tokens are detected within local windows, we additionally construct a window-level knockout policy $\pi_\theta^{\setminus W(v)}$. This policy suppresses all Rock Token candidates appearing in the local high-divergence window $W(v)$ during decoding. The corresponding window-level accuracy delta is:
\begin{equation}
\Delta_{\mathcal{B}}(W(v)) =
\mathrm{Acc}_{\mathcal{B}}\bigl(\pi_\theta^{\setminus W(v)}\bigr)
-
\mathrm{Acc}_{\mathcal{B}}\bigl(\pi_\theta\bigr).
\label{eq:pillar-delta-window}
\end{equation}

A candidate $v$ is a \textbf{Strong Pillar} on $\mathcal{B}$ iff its token-level or window-level perturbation significantly degrades accuracy: (i)
\begin{equation}
\min\{\Delta_{\mathcal{B}}(v), \Delta_{\mathcal{B}}(W(v))\}\le -\varepsilon,
\end{equation}
where $\varepsilon=0.01$, and (ii) a paired bootstrap on per-problem indicators ($10{,}000$ resamples, $\alpha=0.05$) rejects $\mathrm{H}_0\!:\!\Delta=0$ for the corresponding perturbation. Symmetric thresholds at $+\varepsilon$ define \textbf{Strong Stumbling Blocks}, and all other candidates are categorized as Neutral. Accuracies are estimated by sampling at $T{=}1.0$ and averaging over independent rollouts to keep single-trajectory variance below the $\sim\!2\%$ effect scale.

This window-aware criterion serves two purposes. First, it prevents us from over-interpreting isolated high-loss spikes that arise from stylistic mismatch or transient decoding noise. Second, it allows us to distinguish tokens that are individually causal from tokens whose importance only emerges as part of a local reasoning segment. Thus, Pillarhood is treated as a token-in-window causal property rather than a purely token-level artifact.

The screening pool is a strict superset $\widetilde{\mathcal{R}}\supset\mathcal{R}$ of size $K_{\text{screen}}=200$ rather than the definitional $K=100$ set $\mathcal{R}$ of \S\ref{sec:rock_tokens}: $K=100$ is the stability/coverage sweet spot when $\mathcal{R}$ is treated as a vocabulary (qualitative analysis, gradient-zeroing in \S\ref{sec:rq3}), but Pillarhood is a token-in-window causal property and the bootstrap is a search for individual rare-but-real effects, so we trade selection stability below rank $100$ for a wider candidate pool. As a robustness check (Appendix~\ref{app:pillar_multiple_testing}), $6$ of the $10$ Strong Pillars --- including the strongest on each benchmark, ``\,certain'' on MATH-500 and ``-like'' on IFEval --- lie within the $K=100$ core. We probe the post-OPD student on two complementary benchmarks: MATH-500~\cite{lightman2023letsverifystepstep} for the trained reasoning capability and IFEval~\cite{zhou2023instruction} for instruction-following as an out-of-distribution control.

\subsection{Are Rock Tokens pillar? Experiments and findings}
\label{sec:pillar-experiments}

Table~\ref{tab:pillar-census} reports the full Pillar census from the bootstrap procedure across the $|\widetilde{\mathcal{R}}|=200$ screened candidates. The dominant feature of the per-token $\Delta_\mathcal{B}(v)$ distribution is not the tail but the centre: on both benchmarks the great majority of candidates fall inside the $\varepsilon$-band, indicating no detectable causal effect of their removal (visualization in Fig.~\ref{fig:pillar-delta-bar}, Appendix~\ref{app:pillar_multiple_testing}).

\begin{table}[t]
\centering
\caption{Pillar census over the $|\widetilde{\mathcal{R}}|=200$ screened Rock-Token candidates, by benchmark.}
\label{tab:pillar-census}
\begin{tabular}{lcccc}
\toprule
\textbf{Benchmark} & \textbf{Strong Pillar} & \textbf{Neutral} & \textbf{Strong Stumbling} & \textbf{Baseline acc.} \\
\midrule
MATH-500 & 7 (3.5\%)  & 193 (96.5\%) & 0 (0.0\%) & 0.750 \\
IFEval   & 3 (1.5\%)  & 197 (98.5\%) & 0 (0.0\%) & 0.752 \\
\midrule
\multicolumn{4}{l}{Cross-task agreement: 190 N$|$N, 7 P$|$N, 3 N$|$P, \textbf{0} P$|$P} & \\
\bottomrule
\end{tabular}
\end{table}

\paragraph{Multiple-testing considerations.} At $|\widetilde{\mathcal{R}}|=200$ candidates, $\sim\!10$ false rejections are expected by chance at $\alpha=0.05$, and neither Bonferroni nor Benjamini--Hochberg ($q\le 0.20$) retains any individual token (Appendix~\ref{app:pillar_multiple_testing}). We therefore read the per-token identifications above as exploratory \emph{candidate} Pillars. The population-level claim is carried by the \emph{sign} of the rejections: all $10$ fall on the Pillar (negative-$\Delta$) side and zero on the Stumbling side, a $10{:}0$ split with exact probability $2^{-10}\approx 0.002$ under the multiplicity-noise null --- a single global test that does not inflate with the number of candidates screened.

\medskip

Three observations follow. \textbf{First}, Pillars are rare: $3.5\%$ on MATH-500 and $1.5\%$ on IFEval --- the persistent high-KL criterion of \S\ref{ref:recalcitrant_tokens} over-selects causally relevant tokens by more than an order of magnitude. \textbf{Second}, Pillarhood is task-specific: no token is a Strong Pillar on both benchmarks (P$|$P$=0$ in Table~\ref{tab:pillar-census}), so the role of a token type depends on the task. \textbf{Third}, Strong Pillars are content-bearing tokens at decision points, not the structural delimiters that dominate the surface clustering of $\mathcal{R}$ in \S\ref{ref:recalcitrant_tokens}: MATH-500 examples include ``\,certain'', ``\,strategic'', ``\,Initialize'', ``\,programs'', ``Do''; IFEval includes ``\,tools'', ``\,trade'', ``-like''. The symmetric Strong-Stumbling direction is empty under inference-time knockout, an asymmetry we revisit at training time in \S\ref{sec:rq3}.

\subsection{Does Rock Token follow previous findings on `Forking Tokens'?}
\label{sec:pillar-vs-forking}

\emph{Intuition.} The dominant prior framing casts ``important tokens'' as high-entropy decision points where the policy splits between alternative reasoning continuations: forking tokens in RLVR~\cite{wangbeyond}, and the (entropy~$\cup$~student--teacher-divergence) taxonomy formalized for OPD by TIP~\cite{tip2026} and related token-reweighting work~\cite{yang2025token,wu2025mitigating}. If our Pillars were the inference-time projection of any of these training-time importance criteria, knockout $\Delta_\mathcal{B}(v)$ should be predicted by entropy, by KL, or by some combination thereof.

\begin{figure}[t]
  \centering
  \includegraphics[width=\linewidth]{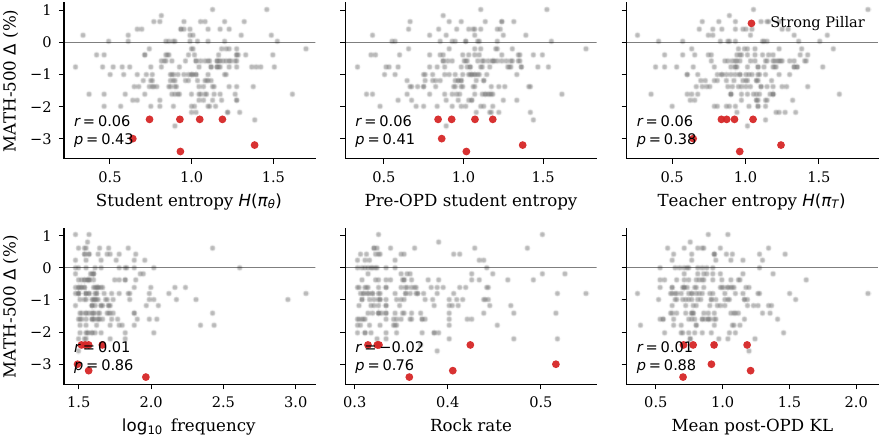}
  \caption{\textbf{Pillarhood is not predicted by entropy, frequency, or
  loss.} MATH-500 knockout $\Delta$ for each of the $|\widetilde{\mathcal{R}}|=200$ screened candidates
  (Strong Pillars in red) plotted against six candidate predictors:
  post- and pre-OPD student entropy, teacher entropy, log-frequency, rock
  rate, and mean post-OPD KL. Annotated $r,p$ are Pearson correlations
  over all 200 candidates. None reach $|r|>0.07$.}
  \label{fig:pillar-vs-predictors}
\end{figure}

\emph{Findings.} Figure~\ref{fig:pillar-vs-predictors} scatters MATH-500 knockout $\Delta_{\mathrm{MATH500}}(v)$ against six candidate predictors --- pre/post-OPD student entropy, teacher entropy, $\log_{10}$ frequency, rock rate, and post-OPD KL --- and \emph{none} reaches $|r|>0.07$; the same conclusion holds on IFEval and on three additional loss-based predictors (Table~\ref{tab:pillar-correlations}, Appendix~\ref{app:pillar_correlations}). Pillars are scattered across the entire entropy range, not concentrated where forking tokens reside, and they are not the tokens with the largest residual KL --- the criterion that $\mathcal{R}$ itself maximizes. Pillarhood is therefore an \emph{orthogonal} cut: a causal, behaviour-defined property that the existing entropy- and divergence-based taxonomies fail to surface, with the immediate methodological corollary that any token-importance scheme keyed to those signals risks suppressing exactly the tokens the student cannot reason without.

\begin{takeaway}
\textcolor{tabblue}{\textbf{Take-away}}~
Pillar Tokens are not Forking Tokens. Their causal role at inference is
uncorrelated with student entropy, teacher entropy, frequency, or
post-OPD KL ($|r| < 0.07$ on MATH-500), placing them outside the
importance taxonomies of prior work~\cite{wangbeyond,tip2026}.
\end{takeaway}

\section{Selection of Ratio}
\label{app:rock_ratio}
Choosing the selection size. The score $R(v)$ provides a ranking; we still need a cutoff $K$ that determines $|\mathcal{R}|$. Two competing criteria bear on this choice. First, $K$ should be small enough that the top-$K$ set is stable — i.e., reproducible across resamplings of the rollout corpus. Second, $K$ should be large enough that $\mathcal{R}$ accounts for a substantial fraction of the corpus-level KL, so that reweighting decisions made on $\mathcal{R}$ have meaningful effect on $\mathcal{L}_{\text{OPD}}$.

\textbf{Procedure} We sweep $K$ jointly with sample size. For each $n \in {50, 100, 200, 300, 400}$, we compute $\widehat R(v)$ from the corresponding sub-corpus and select its top-$K$ tokens; we then measure the Jaccard index of this set against the top-$K$ obtained at $n = 500$. We additionally compute the cumulative coverage $\sum_{v \in \mathcal{R}} \widehat R(v) / \mathcal{L}_{\text{OPD}}^{\text{corpus}}$ at $n = 500$ as $K$ grows.

\textbf{Results} Figure~\ref{fig:rock-id}(b) plots Jaccard against $K$, with one curve per $n$ and an averaged curve. Average Jaccard exceeds $0.70$ for $K \leq 100$ and decays toward a $\approx 0.57$ floor for $K \geq 150$ — beyond which the additional tokens are no longer reliably reproduced across sub-corpora. Figure~\ref{fig:rock-id}(c) overlays the corresponding coverage curve: at $K = 100$, $\mathcal{R}$ accounts for $\approx 60\%$ of the corpus-level KL while remaining within the stable regime. We therefore set $K = 100$ ($\approx 2.5\%$ of the observed vocabulary) as the \emph{definitional} Rock-Token set $\mathcal{R}$ used wherever a stable, reproducible vocabulary is required --- the qualitative cluster analysis below, and the training-time interventions of Section~\ref{sec:rq3}. This choice represents an explicit tradeoff: $K = 50$ would yield modestly higher stability ($J \approx 0.81$) but reduce coverage to $\approx 49\%$. For the per-token causal analysis of Appendix~\ref{sec:pillar-experiments} we instead \emph{screen} a strict superset of size $K_{\text{screen}}=200$: that section's bootstrap is a search for individual rare-but-real Pillars, and the broader candidate pool both increases the chance of detection and lets us probe the regime in which selection stability has decayed below $J\!\approx\!0.6$. The two cutoffs serve different purposes --- the smaller $K$ defines $\mathcal{R}$, the larger one widens the screening window --- and we are explicit in Appendix~\ref{sec:pillar-identification} about which is in force.


\newpage

\end{document}